\documentclass[letterpaper, 10 pt, conference]{ieeeconf}  

\IEEEoverridecommandlockouts                              

\usepackage{graphics} 
\usepackage{epsfig} 
\usepackage{mathptmx} 
\usepackage{times} 
\usepackage{amsmath} 
\usepackage{amssymb}  
\usepackage{cite}
\usepackage{gensymb}
\usepackage{subfigure}
\usepackage{xcolor}

\title{\LARGE \bf
Effective Underwater Glider Path Planning in Dynamic 3D Environments Using Multi-Point Potential Fields
}

\author{Hanzhi Yang$^{1}$ and Nina Mahmoudian$^{1}$
\thanks{$^{1}$Hanzhi Yang and Dr. Nina Mahmoudian are with the school of Mechanical Engineering, Purdue University, West Lafayette, IN, USA. {\tt yang1118, ninam@purdue.edu}}
}
\begin{document}

\maketitle
\thispagestyle{empty}
\pagestyle{empty}

\begin{abstract}
Underwater gliders (UGs) have emerged as highly effective unmanned vehicles for ocean exploration. However, their operation in dynamic and complex underwater environments necessitates robust path-planning strategies. Previous studies have primarily focused on global energy or time-efficient path planning in explored environments, overlooking challenges posed by unpredictable flow conditions and unknown obstacles in varying and dynamic areas like fjords and near-harbor waters. This paper introduces and improves a real-time path planning method, Multi-Point Potential Field (MPPF), tailored for UGs operating in 3D space as they are constrained by buoyancy propulsion and internal actuation. The proposed MPPF method addresses obstacles, flow fields, and local minima, enhancing the efficiency and robustness of UG path planning. A low-cost prototype, the Research Oriented Underwater Glider for Hands-on Investigative Engineering (ROUGHIE), is utilized for validation. Through case studies and simulations, the efficacy of the enhanced MPPF method is demonstrated, highlighting its potential for real-world applications in underwater exploration.
\end{abstract}

\section{INTRODUCTION}\label{Introduction}

Underwater gliders (UGs) have become one of the most effective underwater unmanned vehicles (UUV) for ocean explorations. A UG accomplishes gliding motions by changing its net buoyancy and position of the center of gravity while its wing provides lift force in the water. The working principle of UGs provides long endurance, low energy consumption, and low noise. As many UG models, such as SLOCUM \cite{ref:SLOCUM}and SeaExplorer \cite{ref:SeaExplorer}, have been developed and applied to explore dynamic and complex underwater environments, path planning is one of the most important features to ensure reliable underwater operations. Due to its lack of external actuators and therefore low speed of motion, UG can be significantly affected by environmental disturbances like currents and waves, which are generally modeled as flow fields. In addition, UGs are unable to avoid obstacles quickly due to internal actuation design. While path planning for energy or time efficiency is crucial to take advantage of the inherent capability of the UGs  \cite{ref:PathPlanningOLD1, ref:PathPlanningOLD2, ref:PathPlanningOLD3}, it is important to consider unknown dynamic conditions into account. This is specifically important when the UGs are deployed in areas such as fjords where the detection of the flow field is needed as water may vary from calm to rough seasonally \cite{ref:Fjord1,ref:Fjord2} or in near-harbor regions where detection of obstacles is needed as moving ships and anchors are present.
Hence, to operate in such an environment without knowledge of flow conditions and possible obstacles, a glider needs to utilize real-time path planning methods focusing on local efficiency. There are many real-time path planning methods \cite{ref:PathPlanningAUV1,ref:PathPlanningAUV2,ref:PathPlanningAUV3} proposed for autonomous underwater vehicles (AUVs) with external actuators to enable these vehicles to overcome environmental challenges in horizontal 2D space, however, they are not directly implementable onto UGs as they are traditionally following a certain sawtooth trajectory in the vertical plane. Therefore, there is a need for a specific study in 3D space for UGs exploring confined environments. 

\begin{figure}
    \centering
    \includegraphics[width=1\linewidth]{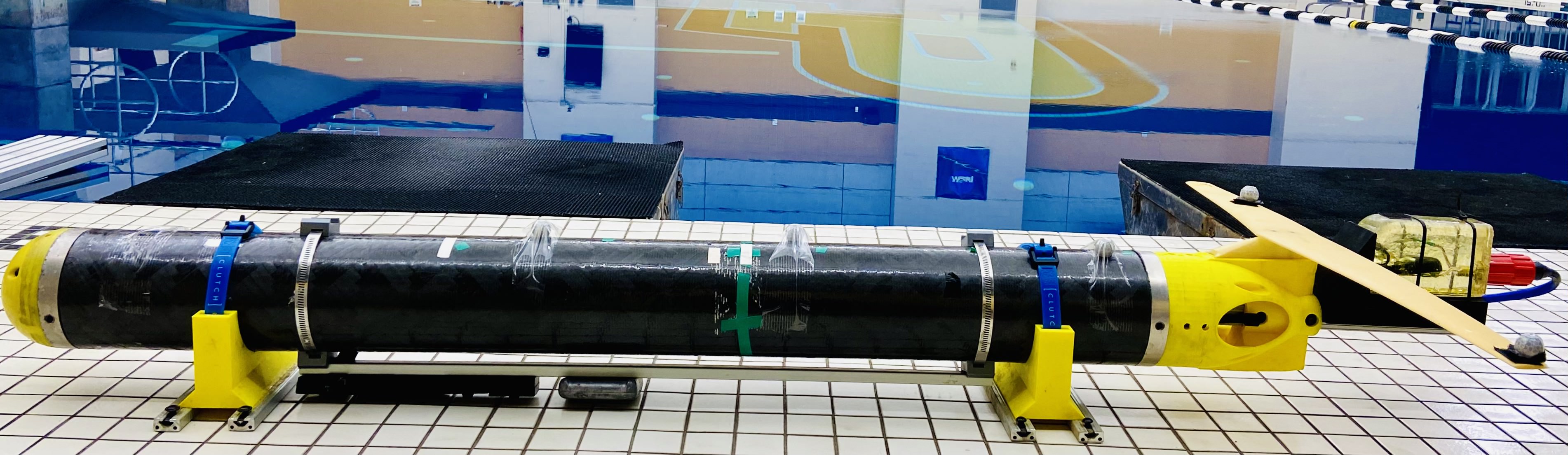}
    \caption{Custom-made underwater glider ROUGHIE was used for validation studies.}
    \label{fig:ROUGHIE}
\end{figure}

The Research Oriented Underwater Glider for Hands-on Investigative Engineering (ROUGHIE) (shown in Fig. \ref{fig:ROUGHIE}) is a low-cost and modular prototype for extending the maneuverability of UGs in constrained spaces like near-shore shallow water areas. It is $1.2 m$ long and weighs $13 kg$. Its small size and weight enable the vehicle shore launch, operation in shallow environments, and indoor pool operations. In a previous work \cite{ref:ROUGHIE}, we implemented a feedforward-PID controller to control pitch, roll, and depth motions. The control system has been validated in a diving pool to show the glider’s high maneuverability in a constrained environment, which makes ROUGHIE able to achieve a relatively small turning radius of $3 m$. With high maneuverability, ROUGHIE can change the heading direction quickly and frequently in a small area, but to deploy the glider in an outdoor area without prior knowledge of potential obstacles and flows, effective real-time path planning is required.

This study builds on 3D space real-time path planning method called Multi-Point Potential Field (MPPF) \cite{ref:MPPF} that works based on Artificial Potential Field (APF). This method was applied to AUV to avoid static and dynamic obstacles in 3D space. To solve the common APF local minima problem, the method detects local minima in a critical zone and commands the AUV to move vertically away from the local minima region. In this paper, we propose a new MPPF method for UGs to explore unknown 3D space with obstacles and flow fields and test the proposed methodology on our glider ROUGHIE. Given a starting point and a target point, the glider first pre-plans a sawtooth trajectory assuming no obstacle or flow would affect it, and then all path planning is proceeded in real-time as the glider follows the pre-planned path. Considering the slow speed of UGs, we add an extra potential field to MPPF that considers the velocities of the moving obstacles. To enhance the time cost of finishing a complete path from the starting point to the target point, the direction and speed of the local flow field around the vehicle are considered in another potential field added to MPPF to make use of the flow field to enhance the local time efficiency. We also examine the ability of the UG to move vertically to stay away from the local minima region. 

This paper is set up as follows: Section \ref{Methodology} reviews the MPPF and proposes an enhanced method with a solution to local minima problem; Section \ref{Case studies} analyzes the method through several application case studies by applying it on ROUGHIE; Section \ref{Results} validates the method by showcasing each case study in simulations; Section \ref{Conclusions} concludes the paper and considers future works.

\section{METHODOLOGY}\label{Methodology}

\subsection{Multi-Point Potential Field (MPPF) review} \label{MMPF review}

\begin{figure}
    \centering
    \includegraphics[width=1\linewidth]{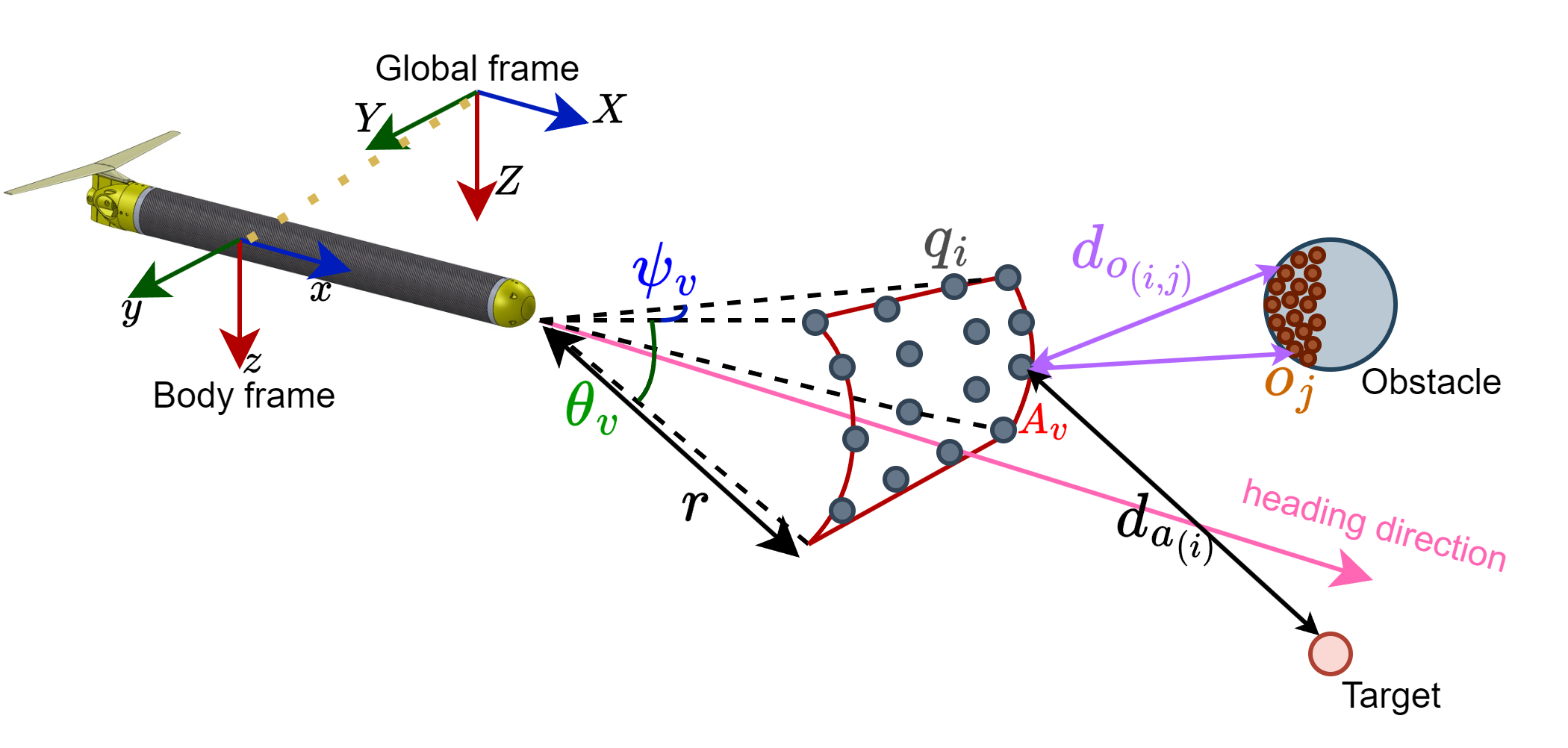}
    \caption{3D illustration of ROUGHIE with MPPF. Attractive potential generates sampling waypoints around the glider and calculates the distance to the target point and repulsive potential generates points on the obstacles and calculates the distance to the points. }
    \label{fig:MPPF}
\end{figure}

The MPPF methodology as shown in Fig. \ref{fig:MPPF} calculates the attraction and repulsion potentials at multiple points sampled on a hemispherical surface ($A_v$) located in front of the vehicle, which is at the current position $P_t$, in its current heading direction for a distance of $r$ and generated based on the vehicle’s maximum turning angles in horizontal ($\psi_m$) and vertical ($\theta_m$) planes and speed. The span of $A_v$ is found by 

\begin{equation}
    \begin{split}
        \psi_v &= \psi \pm \psi_m \\
        \theta_v &= \theta \pm \theta_m
    \end{split}
\end{equation}
where $\psi$ and $\theta$ are the vehicle's current moving angles in horizontal and vertical planes. 

The hemispherical area of interest is discretized into  sampling waypoints in horizontal and vertical directions and the attractive potential at each point is calculated using the distance between the sampling point ($q_i$) and the target point ($q_g$):

\begin{equation}\label{eq:attractive potential}
    U_{attr_{(i)}}(q)=\frac{1}{2}\xi d_{g_{(i)}}^{2}
\end{equation}
in which $d_{g_{(i)}}=|q_g-q_i|$. 

For obstacle avoidance, the obstacles in the detectable range are discretized into  points in horizontal and vertical directions on the surface facing the vehicle. For each waypoint, the distance between it and each sampling point on the surface of obstacles is used to compute the repulsive potential:

\begin{equation}\label{eq:repulsive potential}
    U_{repu_{(i,j)}}(q) = \begin{cases}
            \frac{1}{2}\eta (\frac{1}{d_{o_{(i,j)}}}-\frac{1}{d_t})^2d_{g_{(i)}}^2 & \quad if\, d_{o_{(i,j)}} \leq d_t, \\
            0 & \quad otherwise
        \end{cases}
\end{equation}
in which $d_t$ is the influence distance set for obstacle avoidance safety design, and $d_{o_{(i,j)}}=|q_i-o_j|$ is the distance between the $i-$th sampling waypoint and the $j-$th sampling obstacle point. 

Then the total potential at each sampling waypoint is calculated as a sum of attractive and repulsive potentials: 

\begin{equation}\label{eq:original MPPF}
    U_{tot_{(i)}}(q)=U_{attr_{(i)}}(q) + \sum_{j=1}^N U_{repu_{(i,j)}}(q) 
\end{equation}
in which $N$ is the total number of sampling obstacle points. 

\begin{figure}
    \centering
    \includegraphics[width=1\linewidth]{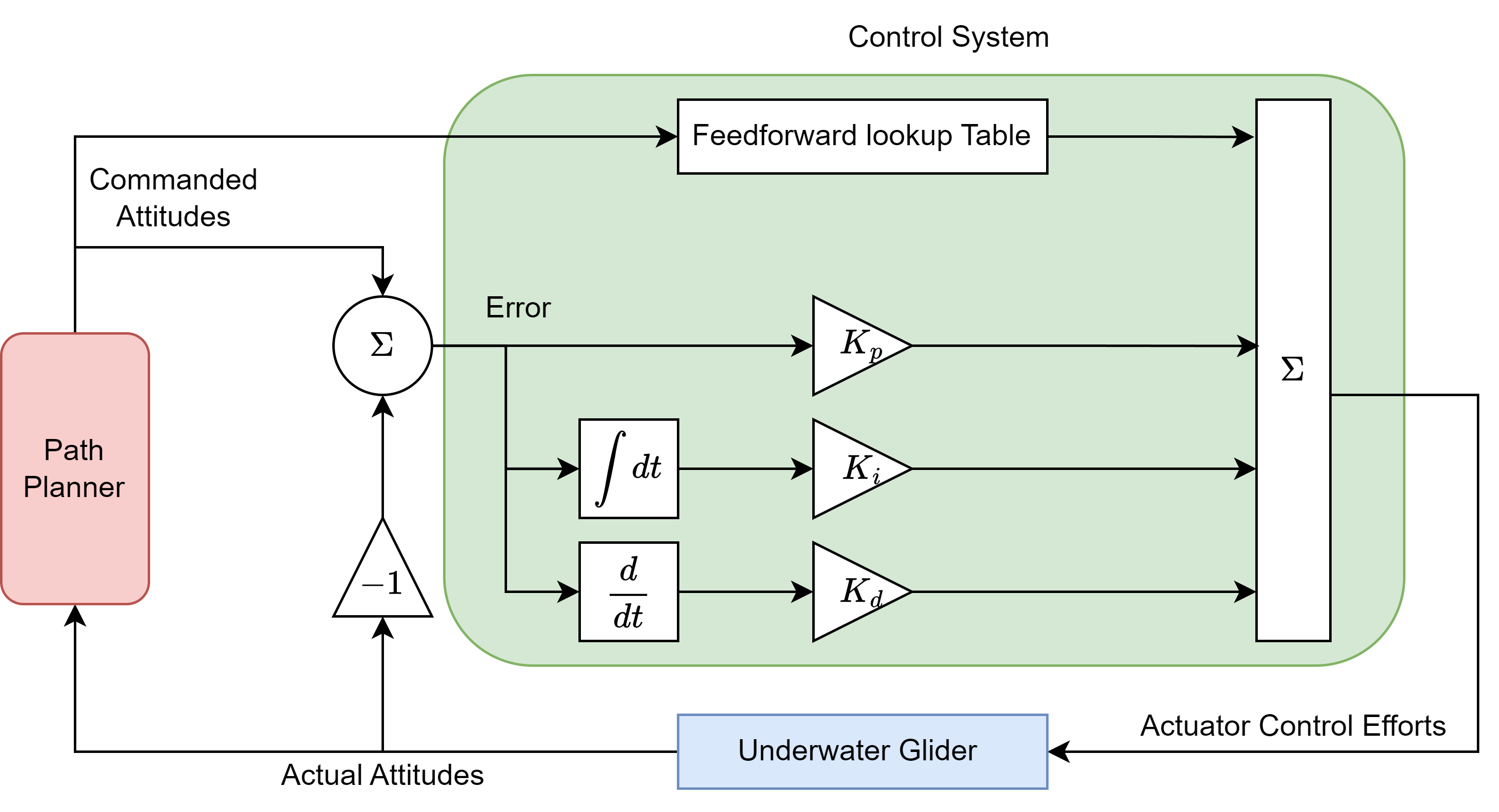}
    \caption{Schematic of closed-loop control system implemented on ROUGHIE. }
    \label{fig:control system}
\end{figure}

The go-to position toward which the vehicle moves for the next step can be determined by finding the waypoint with minimum total potential:
\begin{equation}\label{eq:go-to point}
    P_{t+1} = \arg\min_{q_i \in A_v} (U_{tot})
\end{equation}
The cartesian coordinates of $P_{t+1}$ calculated by MPPF are converted to spherical coordinates consisting of moving angles in horizontal and vertical planes and go-to depth which are then used as command input for the system controller (as shown in Fig. \ref{fig:control system}).

Local minima issues occur when there exists a symmetrical environment or concave-shaped obstacles in the heading direction of the vehicle. In such areas, the vehicle would stop or get trapped in a closed loop because the potential field resultant is zero, failing in planning. To avoid local minima problems, MPPF first detects local minima in a critical zone and then commands the vehicle to move vertically up or down or turn around until it is away from the local minima area. Once no further local minima are detected in the critical zone, the vehicle re-plans the path toward the target point.

\subsection{MPPF for underwater gliders}\label{UG MPPF}
To apply MPPF to a UG, certain modifications are needed because of the relatively low maneuverability, compared to an AUV. The gliding angle and heading angle of a UG are equivalent to turning angles in vertical and horizontal planes. First, as it lacks external actuation, a UG needs to follow a sawtooth path from the start point to the target point. For this sawtooth pre-planning, the environment is assumed to be empty and calm with no obstacle or flow field, and the start and end points are assumed to be separated for a long distance. Given the positions of  the start point ($q_s$) and end point ($q_e$), the glider’s maximum reachable depth ($d_{max}$) and gliding angle ($\theta_{max}$), and the water area’s maximum detectable depth ($Z_{max}$). A sawtooth path between $q_s$ and $q_e$ is calculated for three common cases in gliding trajectory design:

\subsubsection{Case 1 -  start point and end point at different depths}

In this case, a simple straight trajectory is generated by connecting the start and end points with a gliding angle 
\begin{equation}\label{eq:gliding angle}
    \theta_d = \arctan(\frac{z_s-z_e}{\sqrt{(x_e-x_s)^2+(y_e-y_s)^2}})
\end{equation}
and target depth 
\begin{equation}\label{eq:target depth}
    z_d = z_e
\end{equation}

\subsubsection{Case 2 -  start point and end point at the same depth, $Z_{max} \leq d_{max} $}

In this case, a midpoint is set to form a sawtooth trajectory between  start and end points. The position of the midpoint is calculated as
\begin{equation}\label{eq:midpoint}
    q_m = \begin{bmatrix}
            \frac{1}{2}(x_s+x_e)\\
            \frac{1}{2}(y_s+y_e)\\
            Z_{max}-Z_t
        \end{bmatrix}
\end{equation}
in which $Z_t \geq 0$ is a parameter set for safety purpose. The generated sawtooth path in this case consists of two connected segments each one of which is the same as described in case 1, so the gliding angle and target depth for each segment can be found using the Eq. \eqref{eq:gliding angle} and Eq. \eqref{eq:target depth}. 

\subsubsection{Case 3 - start point and end point at the same depth, $Z_{max} > d_{max}$}

In this case, more than one midpoint is needed to form a continuous sawtooth path. First, a maximum horizontal reachable distance is computed as
\begin{equation}
    h_{d_{max}} = 2\cdot\frac{d_{max}}{\frac{1}{2}\theta_{max}}
\end{equation}
with which the first waypoint at the same depth as q\_s and q\_e is located at
\begin{equation}
    q_{w_{(1)}} = q_s + h_{d_{max}} \cdot \frac{q_e-q_s}{\lVert q_e-q_s \rVert}
\end{equation}
then the first midpoint $q_{m_{(1)}}$ between the start point and the first waypoint can be calculated using Eq. \eqref{eq:midpoint}.  Note that half of the maximum reachable gliding angle is used to generate the sawtooth in order to leave enough turning space for trajectory adjustments and obstacle avoidance. After the first waypoint, this process is repeated until the distance between the glider and the end point is less than $h_{d_{max}}$. The depth of the final midpoint is found using the final waypoint $q_{w_{(-1)}}$, the end point $q_e$, and maximum gliding angle $\theta_{max}$
\begin{equation}
    z_{m_{(-1)}} = z_e + \tan(\frac{\theta_{max}}{2})\cdot\frac{\lVert q_e-q_{w_{(-1)}} \rVert}{2}
\end{equation}
and the desired depth and gliding angle can be found using Eq. \eqref{eq:gliding angle} and Eq. \eqref{eq:target depth}. 

The attractive potential field (Eq.\eqref{eq:attractive potential}) is used to follow the pre-planned sawtooth trajectory and track each waypoint while the repulsive potential field (Eq.\eqref{eq:repulsive potential}) is used to avoid obstacles. Since UG is a slow-moving vehicle in water and can not make rapid turns, to avoid dynamic obstacles more efficiently, an extra repulsive potential field involving the velocities of the vehicle and the detected obstacles is added as 
\begin{equation}\label{eq:velocity repulsive}
    U_{repu,v_{(i,j)}}(q) = \begin{cases}
            \frac{1}{2}\tau\frac{V_{UO}}{d_{o_{(i,j)}}} & \quad if\, V_{UO}\geq0 \, and\,  d_{o_{(i,j)}} \leq d_t, \\ 
            0 & \quad otherwise
    \end{cases}
\end{equation}
where $V_{UO} = (V_{UG_{(i)}}-V_{o_{(j)}})^T\frac{o_j-q_i}{\lVert o_j-q_i\rVert}$ is the relative velocity component between the glider and the obstacle point in the direction from sampling waypoint to the obstacle surface point, with $V_{o_{(j)}}$ being the detected velocity of the obstacle surface point $o_j$ and $V_{UG_{(i)}}$ being the velocity of the glider if moving in the direction of $q_i-P_t$. If $V_{UO} \geq 0$, that means the obstacle point is moving toward the glider, and thus certain trajectory adjustment is needed to avoid collision; if $V_{UO}<0$, the obstacle point is moving away from the glider, so the velocity component is not necessary in the potential field. 

Furthermore, because in open water areas, the impact of environmental disturbances such as waves and currents on UG’s motions is significant, another potential field considering the local flow field is added. It is assumed that within a small range near the glider, the flow field is unidirectional with constant speed. The flow field potential is expressed as
\begin{equation}\label{eq:flow field potential}
    U_{flow_{(i)}}(q) = \begin{cases}
            \frac{1}{2}\kappa \lVert V_{flow}-V_{UG_{(i)}} \rVert^2 & \quad if\, |\gamma|\leq\psi_{max}, \\
            \frac{1}{2}\kappa \lVert -V_{flow}-V_{UG_{(i)}} \rVert^2 & \quad if\, |\gamma|\geq\frac{\pi}{2}+\psi_{max}, \\
            0 & \quad otherwise
        \end{cases}
\end{equation}
where $V_{flow}$ is the measured local flow field velocity and $\gamma$ is the angle between $V_{flow}$ and $V_{UG_{(i)}}$. The basic idea of the flow field potential is to make use of the flow in the same direction of UG’s heading to make its motions faster, and if the flow field is of an opposite direction, then move the UG against the flows to avoid larger drifting due to currents from the pre-planned path. 

As a result, with the proposed velocity repulsive potential and flow field potential, the total potential for UG in a dynamic environment yields
\begin{equation}\label{eq:total potential}
    \begin{split}
        U_{tot_{(i)}}(q) & = U_{attr_{(i)}}(q) + \sum_{j=1}^N U_{repu_{(i,j)}}(q) \\
        &+ \sum_{j=1}^N U_{repu,v_{(i,j)}}(q) + U_{flow_{(i)}}(q)
    \end{split}
\end{equation}
and the go-to point is the sampling waypoint with minimum total potential as mentioned in Eq. \eqref{eq:go-to point}.

\subsection{Local minima problem} \label{Local minima problem}

MPPF commands the vehicle to move vertically or turn around when facing local minima issues. UGs are unable to turn around rapidly, and therefore to move away from local minima regions it is commanded to move vertically up to return to gliding maneuvers quickly.  In cases where the glider is too close to the water surface or moving vertically up is not achievable, then it needs to move vertically down. ROUGHIE utilizes a feedforward-PID control system for pitch, roll, and depth control (see Fig. \ref{fig:control system}). To move vertically, zero pitch and roll angles are desired and the desired depth is gradually changed until the local minima region is avoided. 


\section{CASE STUDY}\label{Case studies}

To illustrate the performance of the proposed path planning methodology on underwater gliders, several case studies are conducted using ROUGHIE glider. The glider is $1.2 m$ long and weighs $13 kg$. The maximum gliding angle ($\theta_m$) is $45 \degree$ and the maximum heading angle ($\psi_m$) is $20 \degree$.  For obstacle detection, a sonar echosounder is considered to be installed at the glider’s front nose, with sensing range in horizontal and vertical planes of $100 m$, $120 \degree$, and $30 \degree$. For local flow field measurement, an Acoustic Doppler Current Profiler (ADCP) is added to ROUGHIE. The hemispherical area of interest $A_v$  is located in front of the glider in its heading direction for a distance of $r=V_{UG}\Delta t$. The path planning rate is selected to be $\Delta t = 1$ second and the average gliding speed, due to the asymmetrical dynamics of underwater glider \cite{ref:UGDynamics}, is $V_{UG} = 0.5 m/s$ when heading downward and $V_{UG} = 0.3 m/s$ when heading upward. $A_v$ is discretized into a $5\times5$ sampling waypoint matrix and the obstacle surface, once detected, is discretized into a $3\times3$ sampling point matrix. The potential field parameters are taken as $\xi=0.1$, $\eta=10$, $\kappa=0.1$, $\tau=0.1$, and the influence distance is set as $d_t=2.0\times(R_{obs}+R_{UG})$ where $R_{obs}$ and $R_{UG}$ are the maximum radius of the detected obstacles and the size of the glider.  For local minima detection, the range of the critical zone is set to be $R_{CZ} = R_{obs}+R_{UG}+10r$. 

The simulation is set in a scenario of a crowded near-harbor area with randomly placed obstacles acting as the ships and their anchors. The obstacles are modeled as spherical rigid bodies. The water area for simulation is considered to be a $100\times100\times50m^3$ space while the maximum reachable depth of the glider is $30 m$ beneath the surface. For simulation, the speeds of moving obstacles in the environment and the average speed of the flow field are set to be less than the average speed of the glider. A vortex flow field is considered and modeled as  
\begin{equation}\label{eq:Vortex}
    \begin{split}
        \dot{X} &= -\pi A \sin (\pi \frac{X}{s})\cos (\pi \frac{Y}{s})\frac{Z_{max}-Z}{Z_{max}} \\ 
        \dot{Y} &= \pi A \cos (\pi \frac{X}{s})\sin (\pi \frac{Y}{s})\frac{Z_{max}-Z}{Z_{max}} \\ 
        \dot{Z} &= 0
    \end{split}
\end{equation}
where $A$ is the amplitude of the flow field velocity vector and $s$ is the diameter of each vortex. The velocity of flows decreases as the water goes deeper. The impact of flows acting on the glider is a complex fluid dynamics problem, so in this work, we use a simplified kinematic model to estimate the influence of flows on the glider’s motion performance:  
\begin{equation}
    \dot{P} = V_{UG} + V_{flow}
\end{equation}
in which $P$ denotes the position of ROUGHIE in the global frame and $V_{UG}$ is the vehicle velocity in still water. 

Simulation studies focus on evaluating the effectiveness of the proposed methodology in following sawtooth paths, avoiding static obstacles and local minima, avoiding dynamic obstacles, and gliding in  flow fields. Performances of ROUGHIE utilizing the original MPPF and the proposed version of MPPF methods are compared based on the time cost to finish the entire pre-planned path and the drifting distance from the target point due to the flow field environment.  


\section{SIMULATION RESULTS AND DISCUSSION}\label{Results}

The simulation examines the glider's performance in (1) sawtooth path pre-planning and path following using attractive potential, (2) static obstacle avoidance and local minima solution using repulsive potential and vertically moving control method, (3) dynamic obstacle avoidance using the additional velocity repulsive potential, and (4) flow field efficiency test using the proposed flow field potential.

\subsection{Sawtooth path following} \label{Sawtooth following}

\begin{figure}
    \centering
    \includegraphics[width=1\linewidth]{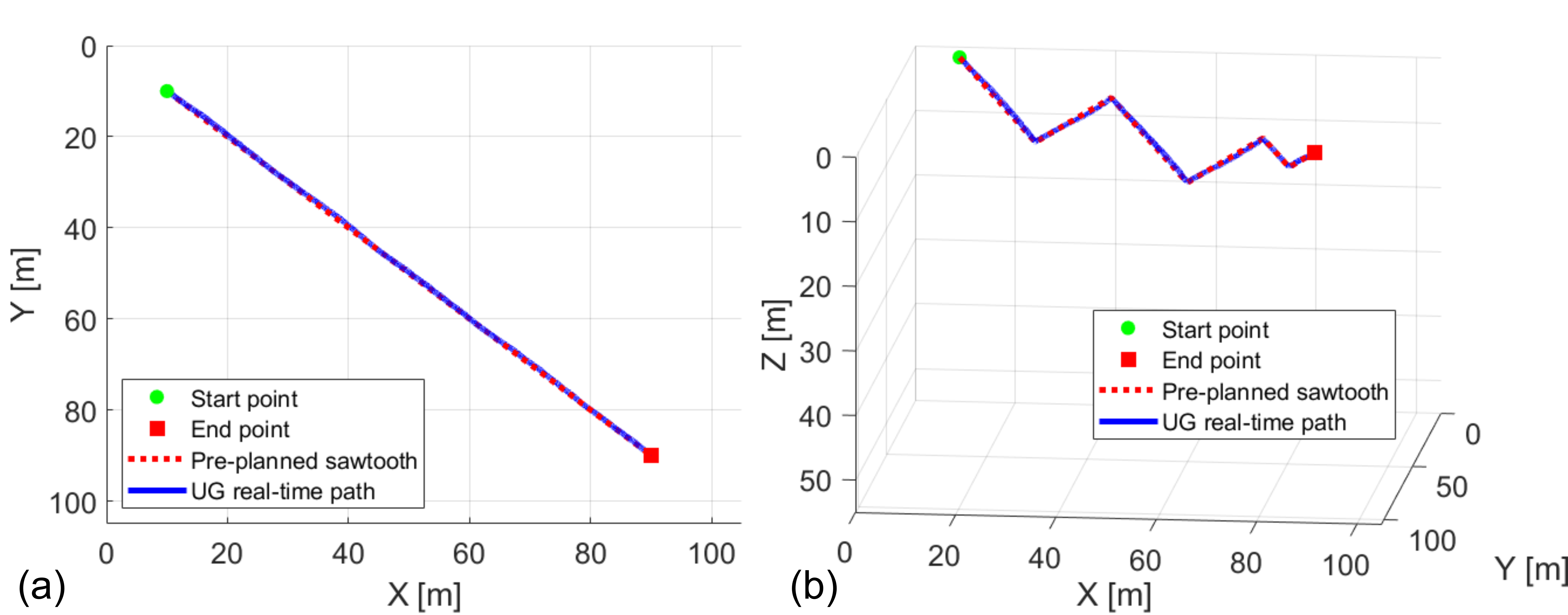}
    \caption{Pre-planned sawtooth trajectory tracking using attractive potential, shown in (a) 2D top view and (b) 3D view. }
    \label{fig:sawtooth tracking}
\end{figure}

To demonstrate sawtooth path pre-planning with multiple midpoints between the start point and the end point, in the simulation for this case, the glider maximum reachable depth is temporarily set as $d_{max} = 10 m$. The start point is located at $(10,10,0)$ and the end point is at $(90,90,0)$ for a scenario in which the glider is commanded to cross a water area diagonally and reach the surface at the other corner. As mentioned in section \ref{UG MPPF}, sawtooth pre-planning first assumes the environment to be free of obstacles and disturbance, and attractive potential is used for tracking the pre-planned trajectories (i.e. $U_{tot} = U_{attr}$). Fig. \ref{fig:sawtooth tracking} shows the waypoints tracking result achieved by attractive potential part of MPPF on UG. The glider is initially heading in the positive direction of the x-axis. This path-following scenario functions as the fundament of the following obstacle avoidance and flow field planning cases.

\subsection{Static obstacles and local minima } \label{Local minima result}

\begin{figure}
    \centering
    \includegraphics[width=1\linewidth]{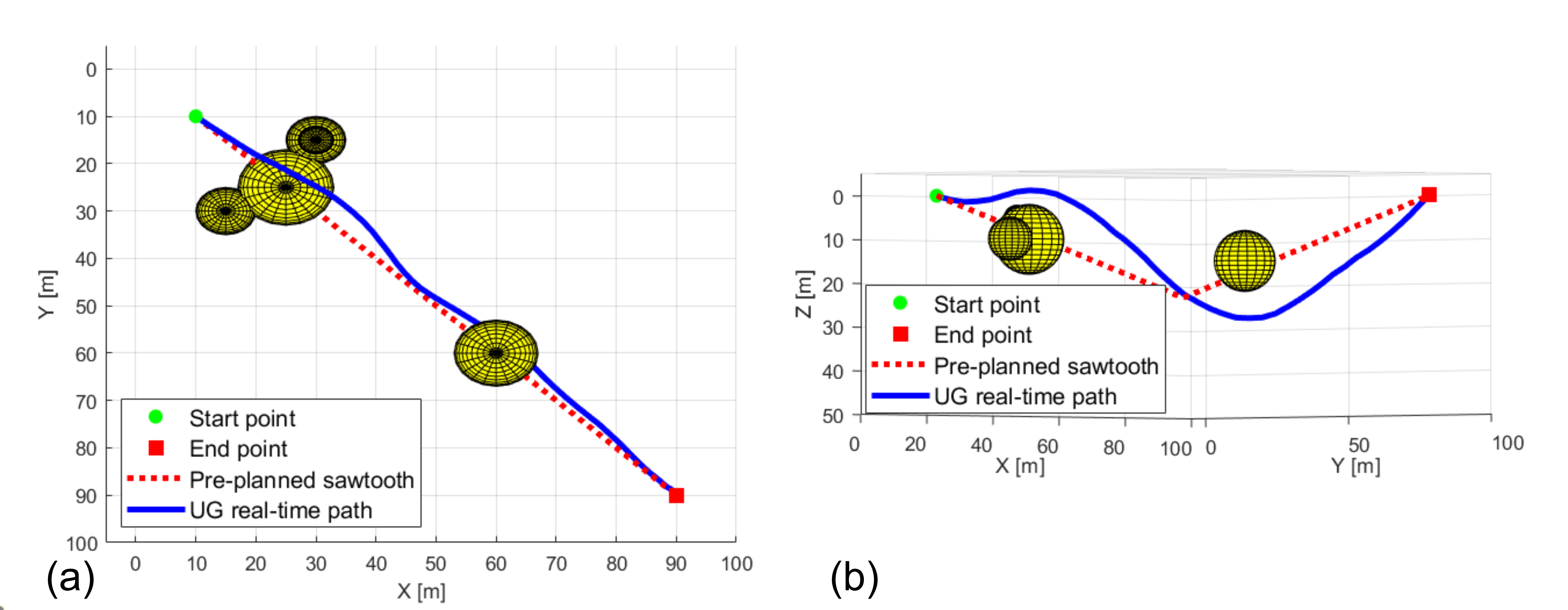}
    \caption{Static obstacle avoidance by repulsive potential, shown in (a) 2D top view and (b) 3D view.}
    \label{fig:static obstacle}
\end{figure}

Fig. \ref{fig:static obstacle} illustrates the glider avoiding static obstacles of various shapes and sizes while following the pre-planned sawtooth trajectory. In this case, the total potential is $U_{tot} = U_{attr} + U_{repu}$. The two obstacles are centered at $(25,25,10)$ and $(60,60,15)$; the first obstacle group is a combination of four spheres with radii of (from left to right then top shown in Fig. \ref{fig:static obstacle}(a)) 5m, 8m, 5m, and 3m, and the second obstacle is a single sphere with a radius of 7m. Because the obstacles are in the middle of the pre-planned sawtooth, the result of the path planner shows a "turn-up" and "turn-down" flight pattern to avoid collision into the spheres. 

\begin{figure}
    \centering
    \includegraphics[width=1\linewidth]{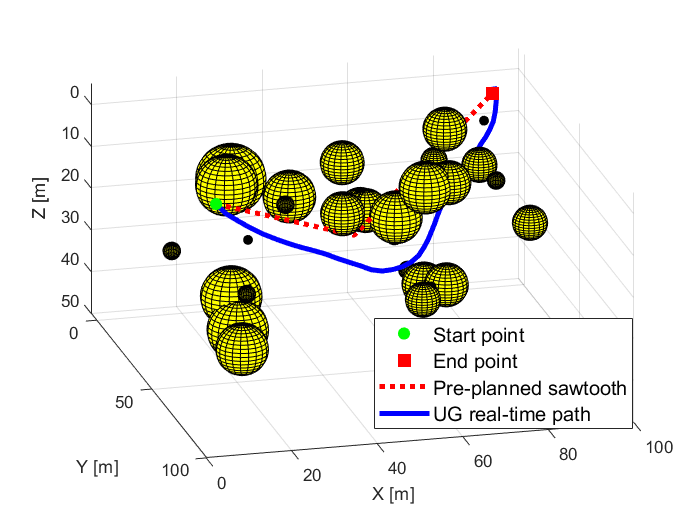}
    \caption{Glider real-time path to avoid randomly generated static obstacles.}
    \label{fig:static obstacle random}
\end{figure}

Fig. \ref{fig:static obstacle random} demonstrates ROUGHIE reaching the target point in a more complex environment full of obstacles of various sizes. The gliding flight starts at $(10,70,0)$ and ends at $(90,10,0)$. Thirty obstacles with radii varying from 0.5 m to 7 m are randomly placed in the environment. As shown, rather than going simply “turn-up” or “turn-down”, the path planner moves the vehicle to go around the obstacles to avoid collision with those blocking its real-time trajectory.  

\begin{figure}
    \centering
    \includegraphics[width=1\linewidth]{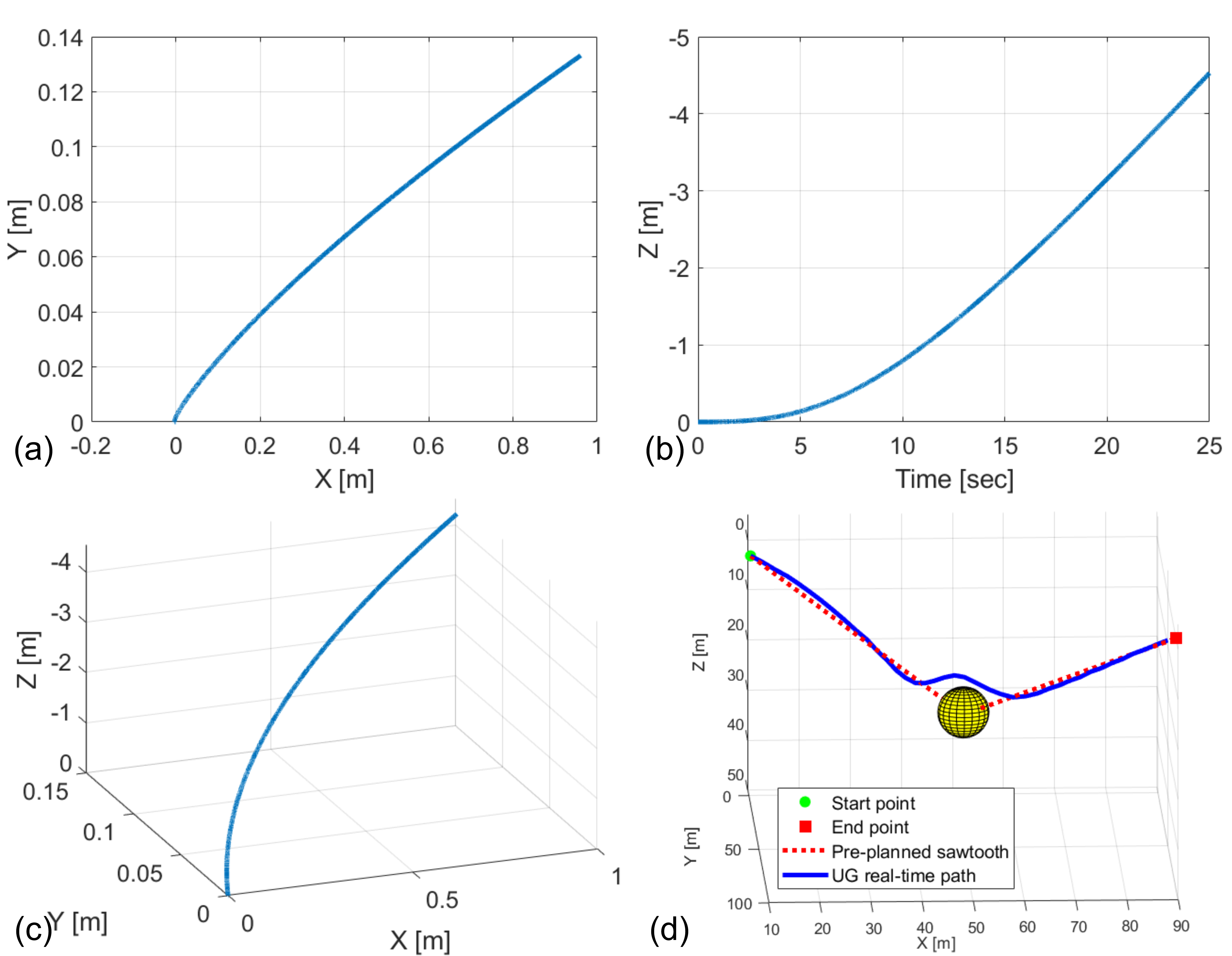}
    \caption{Local minima avoidance trajectory, shown in (a) 2D top view in XY plane,  (b) depth position vs time, (b) 3D view in XYZ space, and (d) a demonstration of avoiding local minima region. }
    \label{fig:local minima}
\end{figure}

Local minima issues occur mostly in the cases of static obstacles where concave-shaped or symmetrical environments are blocking the glider. Fig. \ref{fig:local minima}(c) shows ROUGHIE's capability of moving vertically up (initially heading in positive X direction with zero velocity) to avoid local minima regions using its orientation control system shown in Fig. \ref{fig:control system}. By maintaining near-zero pitch and roll angles, the glider is able to ascend to 4.5 m above its original position in 25 seconds (see Fig. \ref{fig:local minima}(b)). There is a $14 cm$ drifting in the transverse direction and a $1 m$ drifting in the longitudinal axis (see Fig. \ref{fig:local minima}(a)), but compared to the size of the vehicle, obstacles, and the environment, such drifting is negligible.  A demonstration of how the maneuver can be applied to avoid actual local minima area is shown in Fig. \ref{fig:local minima}(d). Note that because the glider's speed is not zero when initiating the local minima avoidance maneuver, the resulting trajectory is not entirely vertical; this further suggests that the critical zone for detecting local minima areas in front of the glider needs to be further than that implemented on a regular AUV.

\subsection{Dynamic obstacles } \label{Obstacle result} 

\begin{figure}
    \centering
    \includegraphics[width=1\linewidth]{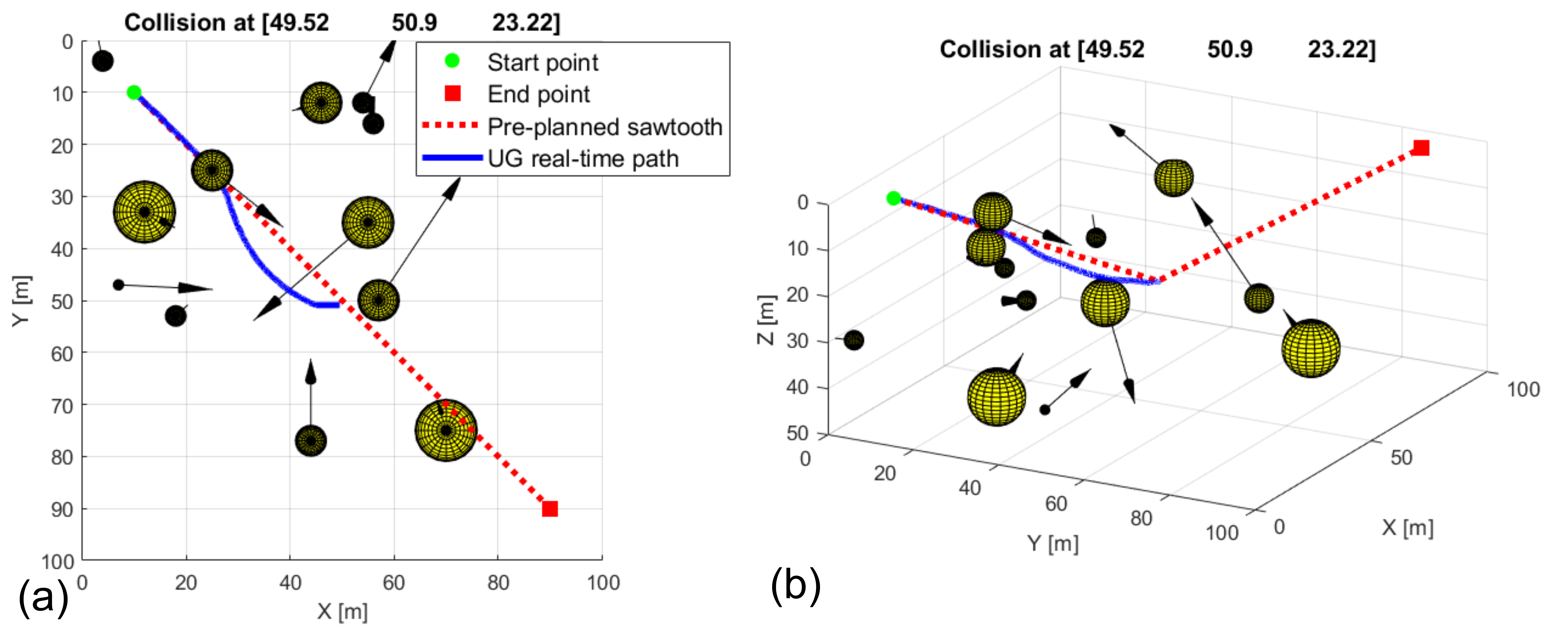}
    \caption{Moving obstacle avoidance using repulsive potential only, shown in (a) 2D top view and (b) 3D view; the arrow on each obstacle indicates the relative amplitude and direction of its velocity.}
    \label{fig:repulsive potential}
\end{figure}

The previous two case studies exemplify the feasibility of adapting the original MPPF to a UG considering the vehicle’s actuation limitations. However, a scenario presented in Fig. \ref{fig:repulsive potential} shows that in a case where random dynamic obstacles are placed in the environment, the original MPPF (as in Eq.\eqref{eq:original MPPF}) may fail to avoid collisions and reach the target point. Twelve spherical obstacles in total are added to the water space. The sizes of these obstacles vary from 0.5 m to 6 m in radius, the maximum moving speed is 0.4 m/s, and the initial positions and headings are randomly generated. The glider crashes with one of the obstacles near the midpoint possibly due to its low speed and constrained turning radius.  

\begin{figure}
    \centering
    \includegraphics[width=1\linewidth]{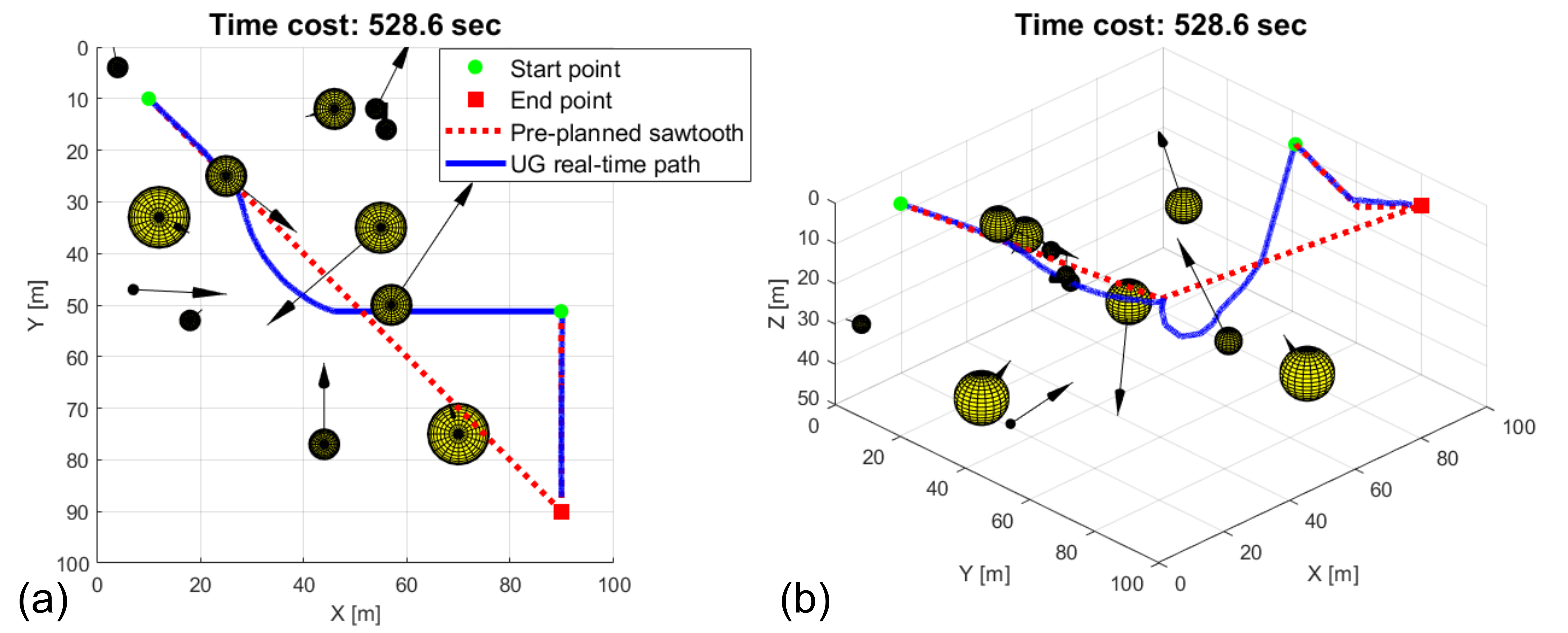}
    \caption{Moving obstacle avoidance using repulsive potential and velocity repulsive potential, shown in (a) 2D top view and (b) 3D view; the arrow on each obstacle indicates the relative amplitude and direction of its velocity.}
    \label{fig:velocity repulsive}
\end{figure}

By applying the proposed velocity potential as Eq.\eqref{eq:velocity repulsive}, successful obstacle avoidance is achieved as shown in Fig. \ref{fig:velocity repulsive}. The environment and obstacle settings are the same as in Fig. \ref{fig:repulsive potential} but the total potential used in this case is $U_{tot} = U_{attr} + U_{repu} + U_{repu,v}$. The glider tends to turn and move in a direction behind the obstacle’s heading to prevent collisions. It is noticeable that the rapid avoidance maneuver eventually causes the vehicle to move in a direction away from the target point, and therefore another sawtooth trajectory is re-planned, following which the glider is able to reach the final target.

\subsection{Flow field } \label{Flow field result}

\begin{figure}
    \centering
    \includegraphics[width=1\linewidth]{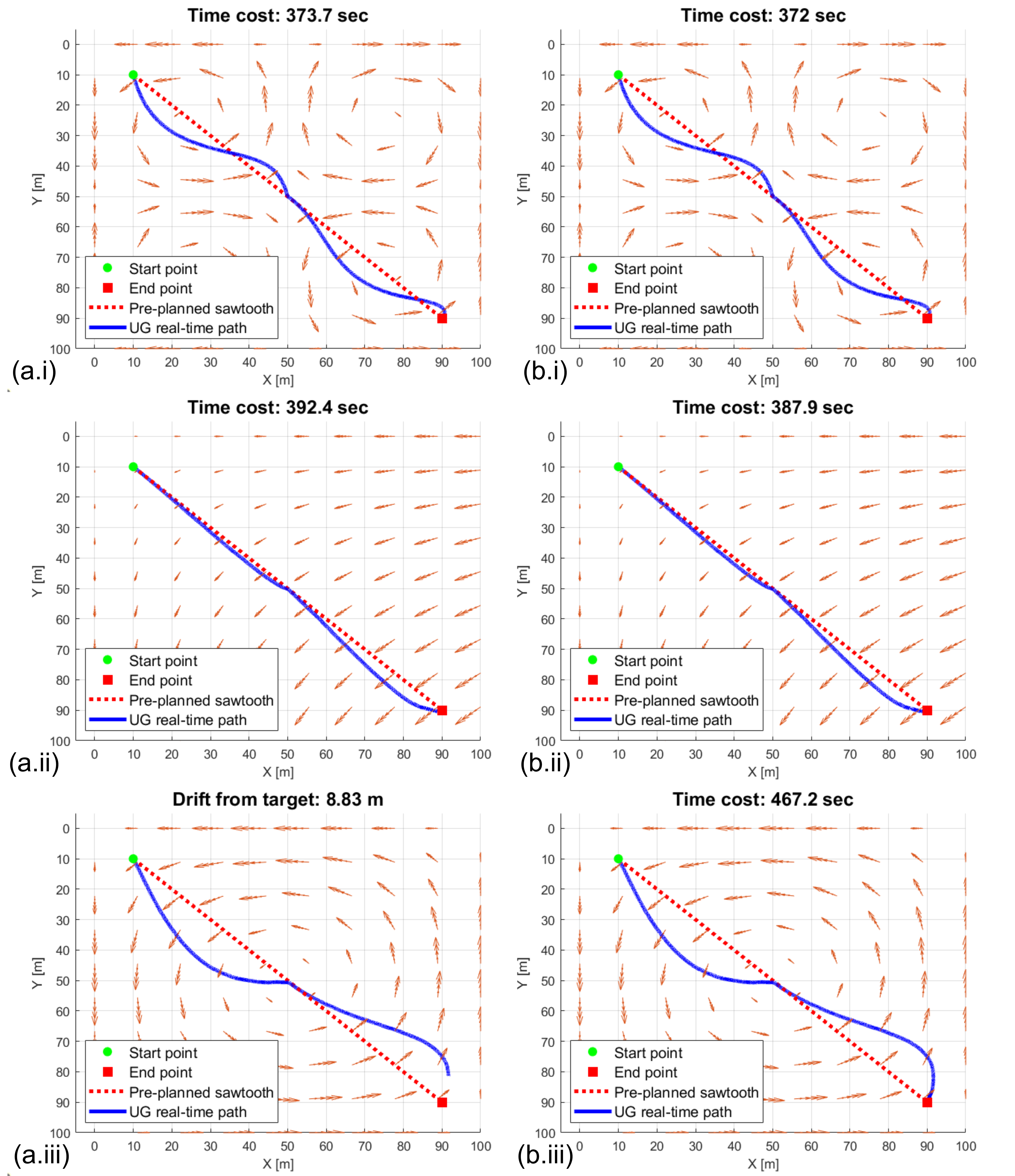}
    \caption{Performance comparisons of path planning in different flow fields by (a) original MPPF and (b) proposed advanced MPPF.}
    \label{fig:flow field}
\end{figure}

\begin{figure}
    \centering
    \includegraphics[width=1\linewidth]{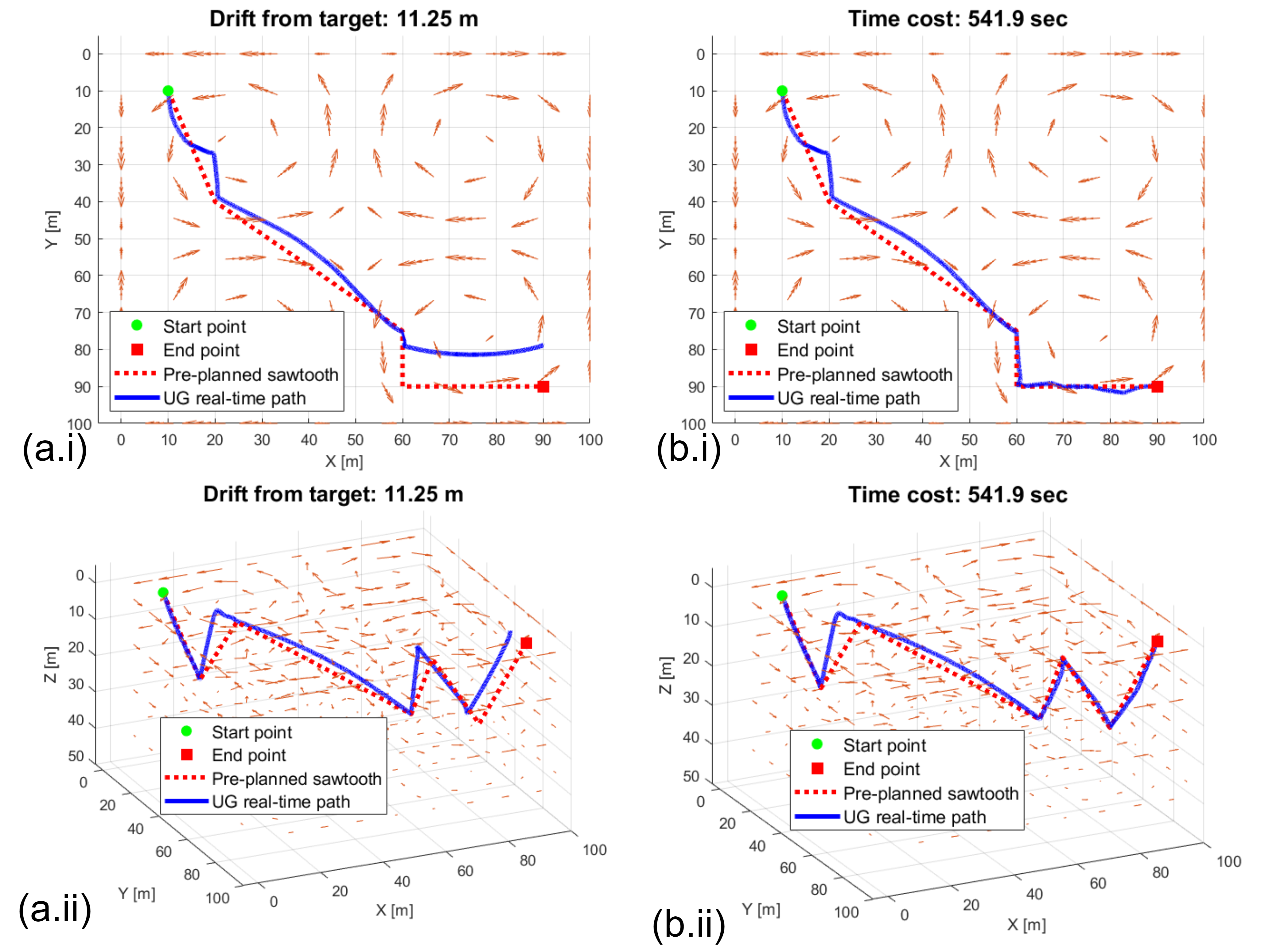}
    \caption{Multiple waypoints following in a flow field environment by (a) original MPPF and (b) proposed method.}
    \label{fig:flow field multi-waypoint}
\end{figure}

\begin{figure}
    \centering
    \includegraphics[width=1\linewidth]{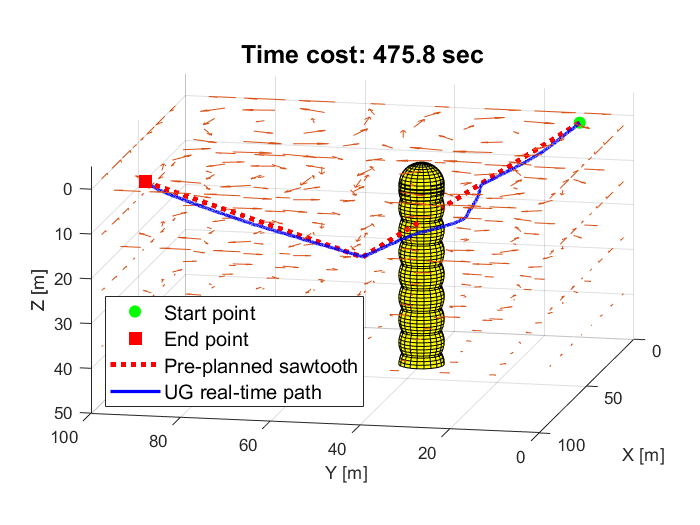}
    \caption{Glider avoiding a cylinder obstacle in a flow field environment.}
    \label{fig:flow field obstacle}
\end{figure}

To showcase the efficacy of the proposed flow field potential, three scenarios are considered: (1) multiple vortices, (2) partial vortex, and (3) single vortex. The vortex environment uses parameters $A=0.1$ and $s = 50$, $1000$, and $100$ for each scenario. As shown in Fig. \ref{fig:flow field}, in the first two scenarios (see Fig. \ref{fig:flow field}(i)(ii)), both original and advanced MPPF result in reaching the target point but advanced MPPF takes less time ($1.7$ seconds lower in time cost for multiple vortices and $4.5$ seconds lower for single vortex) to complete the entire sawtooth trajectory; in the third scenario (see Fig. \ref{fig:flow field}(iii)), original MPPF is unable to reach the target point without further re-planning due to the impact of currents while advanced MPPF can get the glider to the target. 

Note that due to the low speed of UG and pre-planned straight path, the time efficiency improvement is not large compared to the total time cost, so to achieve better time efficiency, a globally planned path is still preferable if the water condition is not completely unknown. Hence, we further test the ability of the proposed method to track multiple preset waypoints which are not necessarily patterned on a straight line. In this case, the glider is commanded, using $U_{tot} = U_{attr}+U_{flow}$, to track 6 waypoints at different depths, resulting in asymmetric gliding angles for its descending and ascending flight segments. As demonstrated in Fig. \ref{fig:flow field multi-waypoint}, the original MPPF leads to an 11.25m drift from the final target point in the flow field environment, but with the additional flow field potential, the vehicle reaches the point with minimal error. This shows that the proposed methodology can be applied to enhance local path efficiency given a globally planned path. 

In addition, in Fig. \ref{fig:flow field obstacle} we illustrate the glider avoiding a cylinder obstacle with a radius of $5m$ blocking the pre-planned sawtooth path while reaching the target point in the flow field environment. The obstacle is centered at $(30,40)$ and blocks the whole vertical space, so the vehicle can not avoid it by moving beneath and above. As shown, utilizing the advanced MPPF as in Eq.\eqref{eq:total potential}, the glider passes around the cylinder while maintaining its trajectory in the flow field environment. This shows that the additional flow field potential does not undermine the capability of avoiding obstacles by the repulsive potential if the glide can overcome the disturbance generated by the water currents.

\section{CONCLUSION AND FUTURE WORK}\label{Conclusions}

In conclusion, this paper presents a novel approach to enhance the real-time path planning of underwater gliders (UGs) in dynamic 3D environments. By adapting and modifying the Multi-Point Potential Field (MPPF) method to suit the unique propulsion mechanism and constraints of UGs, we have demonstrated significant advancements in navigating through environments with obstacle-laden waters and varying flow conditions such as a crowded near-harbor area. Through comprehensive case studies and simulation results conducted on the ROUGHIE prototype, we have validated the efficacy and feasibility of our proposed methodology. Future work will include a software implementation on the ROUGHIE glider and experiments in a real-world water environment.

\bibliographystyle{./IEEEtran} 
\bibliography{./IEEEabrv,./ref}

\end{document}